\documentclass[runningheads,a4paper]{llncs}
\usepackage[T2A,T1]{fontenc}
\usepackage{wrapfig,lipsum,booktabs}
\usepackage[russian,english]{babel}
\usepackage{amssymb,amsfonts,textcomp}
\usepackage{array}
\usepackage{supertabular}
\usepackage{hhline}
\usepackage{hyperref}
\hypersetup{pdftex, colorlinks=true}
\usepackage[pdftex]{graphicx}
\newcommand\textstyleInternetlink[1]{\textcolor[rgb]{0.0,0.0,0.5019608}{#1}}
\date{2015-04-08}
\begin{document}
\title{Determination of the Internet Anonymity Influence on the Level of Aggression and Usage of Obscene Lexis}
\author{Rodmonga Potapova and Denis Gordeev}
\institute{Moscow State Linguistic University, Institute of Applied and Mathematical Linguistics, Moscow, Russia\\
\email{\{RKPotapova, gordeev-d-i\}@yandex.ru}}
\maketitle 
\begin{abstract}This article deals with the analysis of the semantic content of the anonymous Russian-speaking forum 2ch.hk, different verbal means of expressing of the emotional state of aggression are revealed for this site, and aggression is classified by its directions. The lexis of different Russian-and English- speaking anonymous forums (2ch.hk and iichan.hk, 4chan.org) and public community ``MDK'' of the Russian-speaking social network VK is analyzed and compared with the Open Corpus of the Russian language (Opencorpora.org and Brown corpus). The analysis shows that anonymity has no influence on the amount of invective items usage. The effectiveness of moderation was shown for anonymous forums. It was established that Russian obscene lexis was used to express the emotional state of aggression only in 60.4\% of cases for 2ch.hk. These preliminary results show that the Russian obscene lexis on the Internet does not have direct dependence on the emotional state of aggression.
\keywords{imageboard, 2ch.hk, verbal corpus, Internet, aggression, obscene, invective, VK; 4chan.org, anonymity, moderation, nltk, opencorpora}
\end{abstract}
\section{Introduction}
The Internet has connected the entire planet in a global information node. The information is transferred from one end of the planet to another in a fraction of a second. Many of new notions and phenomena are connected with the Internet.
Many of these phenomena are worth of being thoroughly researched. Anonymity shall be mentioned among these phenomena. It is quite possible that anonymity may influence the speech of communicators on the Internet. They may be at ease and venture to act in a way they would have never afforded in their usual verbal communication environment.

Nowadays semantic field aggression in mass media texts is widening and the perception criticality threshold is reducing'' [9]. So questions arise, whether anonymity has any influence on the users' behavior and how much communication standards are affected by anonymous environment. The solution of this issue is directly connected with the development of the Internet in Russia.

This study is focused on anonymous forums, usually called ``imageboards'' because it is possible to attach images to messages there (however, there are text-only imageboards), which are popular nowadays [1]. This study aims at identifying the {\textquotedbl}verbal peculiarity of social networking discourse generation and operation in the world of the electronic mass media environment (based on anonymous forums). R. K. Potapova defines the social networking discourse as ``a distant, indirect, multivector, multidirectional, simultaneous or non-simultaneous electronic macropolylogue, which reflects interpersonal, inter-ethnic, inter-religious, socio-economic, geopolitical, and etc. relationships which influence the usage of specific verbal and paraverbal correlates of written and oral texts`` [10]. A special role in the Russian segment of the Internet belongs to different kinds of invectives, so special attention is paid to their study.

Many works have been devoted to the study of aggression on the Internet. However, many of them were dedicated to Internet addiction [11], bullying [4]. In addition, there are studies devoted to partially anonymous resources, such as
reddit.com [3] where users have accounts and history of their commentaries.
\section{Methods and material of the research}
In the course of this study texts were analyzed from the Russian-speaking anonymous forums 2ch.hk and iichan.hk and public community MDK of the social network VK, moreover, the most popular English-speaking imageboard 4chan.org was put under analysis. This step of the research was aimed at determining the effects of anonymity on the use of obscene lexis and invectives. These websites were selected because of their popularity and different approaches to anonymity (imageboards are anonymous, while MDK is not, iichan is also moderated.)

114,275 messages from 2ch.hk were analyzed, as well as 29134 posts from the site iichan.hk and 3556 messages from the community MDK. This research was based on the method of content analysis. {\textquotedbl}The technique of content analysis makes it possible to obtain an objective and systematic estimation of quantitative content of communicative units{\textquotedbl} [8]. Obscene lexis was used as the marker of the concept of {\textquotedbl}aggression{\textquotedbl}. Although it is not always a sign of aggression, it may affect the communicational culture negatively, as evidenced by common practice and ban of its use on many resources. The programming language Python (version 2.7) was used to count words and to parse messages. In addition, the library NLTK [5] was used in the course of analysis. The frequency analysis was also conducted for the occurrences of Russian-speaking anonymous forums and public community MDK vocabulary in the ``Open corpus of the Russian language'' [7]. The Brown corpus [6] was used to count frequencies for 4chan.org.

2,656 posts from 35 2ch.hk threads were analyzed to determine the degree of aggression in messages. All data were collected in January 2015. The analysis was conducted by a thorough reading of each message by a Russian native speaker to determine the presence of any aggressive component and profanity. If a message contained an aggressive component, the direction of the aggression was specified. All the messages were classified into some target groups (cultural, ethnic etc.)
\subsection{Determination of the effects of anonymity on the usage of the obscene lexis}
During the analysis messages from 3 imageboards and one public community of the Russian social network were analyzed (Table 1).
\begin{table}
\caption{Data obtained during the analysis of anonymous forums and MDK\textbf{ }community lexis}
\begin{center}
\begin{tabular}{|m{0.25in}|m{0.8in}|m{1.1in}|m{1.2in}|m{1.3in}|}
\hline
{ {\textnumero}{\textnumero}} &
{ Internet source} &
{ \textrm{Word count values}} &
{ \textrm{Count values of non-stopword }\textrm{wordforms}} &
{ \textrm{Average count values of words per }\textrm{utterance}}\\\hline
{ 1} & { 2ch.hk} &     { 2648077} & { 177531} & { 23.17}\\\hline
{ 2} & { iichan.hk} &  { 1173309} & { 95500}  & { 40.27}\\\hline
{ 3} & { vk.com/mdk} & { 29476}   & { 6486}   & { 8.29}\\\hline
{ 4} & { 4chan.org} &  { 2109712} & { 70287}  & { 15.09}\\\hline
\end{tabular}
\end{center}
\end{table}

The number of words per message is much smaller for the MDK community, but it is dictated by the communication performed in the form of commentaries to any article. Messages from the moderated forum iichan.hk are almost twice longer than those of 2ch.hk on average. For this reason, it is possible to suggest a hypothesis that efficient moderation may increase the quality of communication; however, the length of a message may not correspond to the amount of transferred information. It is impossible to compare 4chan.org with Russian-speaking forums by this criterion, since the concept of words differs in various languages.

All the semantic items (words) in the forums have been counted by their absolute frequency, stopwords were excluded (Table 2). The array of wordforms, excluding functional words, will be referred to as meaningful words.
\begin{table}
\caption{Most common word forms (types)\protect\footnotemark}
\begin{center}
\begin{supertabular}{|m{0.3in}|m{1.1in}|m{1.1in}|m{1.3in}|m{0.3in}|}
\hline
{ {\textnumero}{\textnumero}} &
\multicolumn{4}{m{4.2in}|}{\centering{ Internet sources}}\\\hline
 &
{ 2ch.hk} &
{ iichan.hk} &
{ vk.com/mdk} &
{ 4chan.org}\\\hhline{~----}
{ 1} & { sage} &
{ \textcyrillic{{\cyrerev}{\cyrt}{\cyro} (this)}} &
{ \textcyrillic{{\cyrerev}{\cyrt}{\cyro} (this)}} &
{ like}\\\hline
{ 2} &
{ \textcyrillic{{\cyrerev}{\cyrt}{\cyro} (this)}} &
{ \textcyrillic{{\cyrv}{\cyrs}{\cyryo} (all)}} &
{ \textcyrillic{{\cyrp}{\cyro}{\cyrd}{\cyra}{\cyrr}{\cyro}{\cyrk} (present)}} &
{ get}\\\hline
{ 3} &
{ \textcyrillic{{\cyrp}{\cyrr}{\cyro}{\cyrs}{\cyrt}{\cyro} (just)}} &
{ \textcyrillic{{\cyre}{\cyrshch}{\cyryo} (more)}} &
{ \textcyrillic{{\cyrt}{\cyre}{\cyrb}{\cyre} (you, Dat.)}} &
{ one}\\\hline

\end{supertabular}
\end{center}
\end{table}
\footnotetext{Words marked by ``*'' are obscene, ``Eng.'' means transliteration of English words, $\sim$ sign shows similar English words.}

The most frequently used wordforms for the imageboard 2ch.hk is ``sage'' and its derivatives ``\textcyrillic{{\cyrs}{\cyra}{\cyrzh}{\cyra}'' (sage, Nom.), ``{\cyrs}{\cyra}{\cyrzh}{\cyri}'' (sage, Gen.)}. This word is a command for website engine, and should be typed in the e-mail field, but users often write it in the message field. Thus, users show their disagreement with the topic of the discussion or with other opinions.

The iichan.hk website is moderated and users are often banned for violation of the site rules. Moreover, there exists an automatic obscene lexis filter aimed at blocking publication of messages containing obscene words. That is why common Russian words are typical for this forum.

When MDK community texts were collected, users created a thread where they exchanged vk.com electronic presents. Trying to attract attention they wrote a large amount of same messages often consisting of repetitions of the same word. That has influenced the statistical data. 

An analysis was then conducted to determine which bigrams do not occur in the ``Open corpus of the Russian language'' for Russian-speaking sites and in the Brown corpus for 4chan.org. ``Opencorpora.org'' [7] consists of Wikipedia and journalist articles, fiction texts and blogs; that is why it contains a very limited volume of colloquial language. The Brown corpus is considered long outdated nowadays, but modern corpora may contain discussions about 4chan.org or obscene lexis. So the Brown corpus is considered sufficient for this research aims. The bigram results are shown in Table 3.
\begin{table}
\caption{Top 5 bigrams not contained in the Opencorpora.org and the Brown corpus}
\begin{center}
\begin{supertabular}{|m{0.2in}|m{1.3in}|m{1.3in}|m{1.3in}|m{0.6in}|}
\hline
{ {\textnumero}{\textnumero}} &
\multicolumn{4}{m{4.6in}|}{\centering{ Internet sources}}\\\hline
 &
2ch.hk &
Iichan.hk &
vk.com/mdk &
4chan.org\\\hhline{~----}
1 &
sage sage&
{ \textcyrillic{{\cyrv}{\cyrs}{\cyryo} {\cyrr}{\cyra}{\cyrv}{\cyrn}{\cyro} (it doesn't matter)}}
&
{ \textcyrillic{{\cyrf}{\cyra}{\cyrs}{\cyrt}{\cyrf}{\cyra}{\cyrs}{\cyrt}
{\cyrf}{\cyra}{\cyrs}{\cyrt} (Eng. fastfast fast)}} &
{ global rule}\\\hline
{ 2} &
{ \textcyrillic{{\cyrs}{\cyra}{\cyrzh}{\cyra} {\cyrs}{\cyra}{\cyrzh}{\cyra} (sage sage)}} &
{ \textcyrillic{{\cyrv}{\cyrs}{\cyryo} {\cyrerev}{\cyrt}{\cyro} (all this)}} &
{ \textcyrillic{{\cyrf}{\cyra}{\cyrs}{\cyrt}
{\cyrf}{\cyra}{\cyrs}{\cyrt}{\cyrf}{\cyra}{\cyrs}{\cyrt} (Eng. fast fastfast)}} &
rule global\\\hline
3 &
{ \textcyrillic{{\cyrs}{\cyra}{\cyrzh}{\cyri} {\cyrs}{\cyra}{\cyrzh}{\cyri} (sage Gen. sage
Gen.)}} &
{ \textcyrillic{{\cyrerev}{\cyrt}{\cyro} {\cyrv}{\cyrs}{\cyryo} (all this)}} &
{ \textcyrillic{{\cyro}{\cyrb}{\cyrm}{\cyre}{\cyrn} {\cyro}{\cyrb}{\cyrm}{\cyre}{\cyrn}
(exchange exchange)}} &
{ desu desu}\\\hline
{ 4} &
{ \foreignlanguage{russian}{{\cyrp}{\cyro}{\cyrsh}{\cyre}{\cyrl} {\cyrn}{\cyra}***
(}f\foreignlanguage{russian}{*}ck\foreignlanguage{russian}{ }off\foreignlanguage{russian}{)}} &
{ \textcyrillic{{\cyrs}{\cyru}{\cyrd}{\cyrya} {\cyrp}{\cyro} {\cyrv}{\cyrs}{\cyre}{\cyrm}{\cyru}
(to all appearances)}} &
{ \textcyrillic{{\cyrp}{\cyro}{\cyrd}{\cyra}{\cyrr}{\cyri}{\cyrt}{\cyre}
{\cyrp}{\cyro}{\cyrd}{\cyra}{\cyrr}{\cyro}{\cyrk} (give me a gift)}} &
{ c*nt c*nt}\\\hline
{ 5} &
{ \textcyrillic{{\cyrm}{\cyro}{\cyrzh}{\cyre}{\cyrsh}{\cyrsftsn}
{\cyrs}{\cyrd}{\cyre}{\cyrl}{\cyra}{\cyrt}{\cyrsftsn} (can do)}} &
{ \textcyrillic{{\cyrv}{\cyrs}{\cyryo} {\cyre}{\cyrshch}{\cyryo} (still)}} &
{ \textcyrillic{{\cyrb}{\cyre}{\cyrs}{\cyrp}{\cyrl}{\cyra}{\cyrt}{\cyrn}{\cyrery}{\cyrishrt}
}\ \textcyrillic{{\cyrp}{\cyro}{\cyrd}{\cyra}{\cyrr}{\cyro}{\cyrk} (free gift)}} &
{ puddi puddi}\\\hline
\end{supertabular}
\end{center}
\end{table} 

Russian obscene collocations and repetitions of the word ``sage'' are typical for 2ch.hk, and they often turn out to be a sign of aggression for this site. Common Russian collocations are typical of iichan.hk. This forum is anime- and Japanese culture-oriented, and indeed users often use Japanese proper names however, they are not frequent at the collocation level. 4chan.org is famous for its jokes, memes and profanity, and indeed they are often mentioned. Such collocations as ``puddi puddi'' (a phrase from a viral Japanese 
pudding advertisement), insulting ``c*nt c*nt'' or local templates (pic related) are commonly used. Finally, an analysis was conducted to find word forms that cannot be found in ``Opencorpora.org'' and the Brown corpus. Then these words were distributed by their absolute frequency (Table 4).
\begin{table}
\caption{Quantitative distribution of word forms missing from ``Opencorpora''}
\begin{flushleft}
\begin{supertabular}{|m{0.3in}|m{1.4in}|m{1.4in}|m{1in}|m{0.7in}|}
\hline
{ {\textnumero\textnumero}} &
\multicolumn{4}{m{4.6in}|}{\centering{ Internet sources}}\\\hline
 &
{ 2ch.hk} &
{ Iichan.hk} &
{ vk.com/mdk} &
{ 4chan.org}\\\hhline{~----}
{ 1} &
{ sage} &
{ \textcyrillic{{\CYRK}{\cyri}{\cyrl}{\cyri} (Kili)}} &
{ \textcyrillic{{\cyrf}{\cyra}{\cyrs}{\cyrt} (fast)}} &
{ f*ggot}\\\hline
{ 2} &
{ \textcyrillic{{\cyrs}{\cyra}{\cyrzh}{\cyri} (sage, Gen.)}} &
{ \textcyrillic{{\CYRR}{\cyre}{\cyrishrt}{\cyrm}{\cyru} (Reimu)}} &
{ \textcyrillic{{\cyrf}{\cyra}{\cyrs}{\cyrt}{\cyrf}{\cyra}{\cyrs}{\cyrt} (fastfast)}} &
{ anon}\\\hline
{ 3} &
{ \textcyrillic{{\cyrt}{\cyrya}{\cyrn} (Jap. -chan)}} &
{ \textcyrillic{{\CYRM}{\cyra}{\cyrr}{\cyri}{\cyrs}{\cyra} (Marisa)}} &
{ \textcyrillic{{\cyrm}{\cyrd}{\cyrk} (mdk)}} &
{ gon}\\\hline
{ 4} &
{ \textcyrillic{{\cyrs}{\cyrf}{\cyri}*{\cyrk}{\cyrt}{\cyre}{\cyrr} (sph*ncter)}} &
{ \textcyrillic{{\CYRK}{\cyre}{\cyrishrt}{\cyrn}{\cyre} (Keyne)}} &
{ \textcyrillic{{\cyrp}{\cyrl}{\cyri}{\cyrz} (Eng. please)}} &
{ f*cked}\\\hline
{ 5} &
{ \textcyrillic{{\cyra}{\cyrn}{\cyro}{\cyrn} (anon)}} &
{ \textcyrillic{{\CYRP}{\cyra}{\cyrch}{\cyru}{\cyrl}{\cyri} (Patchouli)}} &
{ \textcyrillic{{\cyra}{\cyrd}{\cyrm}{\cyri}{\cyrn}{\cyra} (admin Prep. case)}} &
{ f*ggots}\\\hline
\end{supertabular}
\end{flushleft}
\end{table}
The search of words missing from ``Opencorpora'' and the Brown corpus helped to discover words typical for these sites; however, it was associated with some problems. E.g. the word ``\textcyrillic{{\cyrs}{\cyra}{\cyrzh}{\cyra}'' (Eng. sage) was excluded from the set, as it is spelled just the same as the common Russian word. This problem cannot be solved without a sufficient context (that is why in the bigrams list the phrase ``{\cyrs}{\cyra}{\cyrzh}{\cyra} {\cyrs}{\cyra}{\cyrzh}{\cyra}'' (``sage sage'') remained, as this expression is very unusual for Russian, however, other expressions with this word might had been excluded, too)}. However, slang is not very popular on iichan.hk; communications are held using common Russian words, so the lexis of the site intersected considerably with the corpus. For this reason, proper names are mostly presented in this table, which are usually anime or Touhou project (a Japanese scrolling shooter game) characters.

Social networks lexis is common for the MDK community; however, some euphemisms are used here(``\textcyrillic{{\cyrz}{\cyrb}{\cyrs}'', ``fckng good''). The word ``{\cyrm}{\cyra}{\cyrm}{\cyrk}{\cyra}'' (``mother'', diminutive) does not carry any negative sense here, and is used just in the same context as other words with this root. Although diminutive suffices often carry insulting in Russian nowadays.}

Abbreviations are typical of 4chan just as of other English-speaking sites (e.g. ``mfw'' - my face when). In the Russian Internet slang abbreviations are not very common. Obscene words (``f*ggot'', ``f*cked'') and words describing the forum and its users (``4chan'', ``anon'') are also quite popular.

An analysis was also conducted of the obscene lexis percentage on these sites, which made it possible to detect the cultural level of discussions and to suppose if these sites' users are more prone to aggression and if anonymity influences it (Table 5).
\begin{table}
\caption{Statistics of obscene words usage}
\begin{flushleft}
\tablefirsthead{}
\tablehead{}
\tabletail{}
\tablelasttail{}
\begin{supertabular}{|m{0.3in}|m{1.2in}|m{2.5in}|}
\hline
{ {\textnumero}{\textnumero}} &
{ Internet sources} &
Percentage of all significant word forms (\%)\\\hline
{ 1} &
{ 4chan.org} &
{ 3.55}\\\hline
{ 2} &
{ vk.com/mdk} &
{ 1.37}\\\hline
{ 3} &
{ 2ch.hk } &
{ 0.81 }\\\hline
{ 4} &
{ iichan.hk } &
{ 0.01 }\\\hline
\end{supertabular}
\end{flushleft}
\end{table}
Thus, according to the data, anonymity does not influence the percentage of obscene words. Moreover, the amount
of profanity is even bigger for non-anonymous vk.com (although these values fluctuate with the growth of the set).
\subsection{Determination of the relation between the usage of obscene words and the aggression level in messages from
Russian anonymous forum 2ch.hk}
2656 messages from Russian anonymous imageboard 2ch.hk were analyzed to determine the connection between the level of aggression and the use of profanity. The majority (n = 2105) of messages analyzed during the experiment did not contain either signs of aggression or profanity. 20.7\% (n = 551) of messages contained obscene words and/or
were aimed at other users or social groups, which might affect the communication on the forum. 218 messages contained profanity, but it was used to express positive emotions, or for other non-aggressive purposes. 333 messages had signs
of aggression.


Messages containing aggression were classified into 10 groups, according to the target of the aggression (Table 6).
\begin{table}
\caption{Ratio of messages containing aggression, obscene words and neutral utterances.}
\begin{flushleft}
\begin{tabular}{|m{0.4in}|m{2in}|m{1.2in}|}
\hline
{ {\textnumero}{\textnumero}} &
{ Types of aggression} &
{ Count values (\%)}\\\hline
{ 1} &
{ psycho-physiological} &
{ 162 (49)}\\\hline
{ 2} &
{ ethnic} &
{ 70 (21)}\\\hline
{ 3} &
{ political} &
{ 61 (18)}\\\hline
{ 4} &
{ culturological} &
{ 21 (6)}\\\hline
{ 5} &
{ socio-economic } &
{ 13 (4) }\\\hline
{ 6} &
{ geopolitical } &
{ 4 (1)}\\\hline
{ 7} &
{ pragmatic} &
{ 2 (1)}\\\hline
{ 8} &
{ confessional } &
{ 0 (0)}\\\hline
{ 10} &
{ Total} &
{ 333 (100)}\\\hline
\end{tabular}
\end{flushleft}
\end{table}
In most cases (49\%) aggression was expressed towards other communication partners, whose imaginary or real deficiencies such as age, sexual orientation or state of mind were highlighted. In 21\% of cases (70 messages) aggression was expressed towards a certain nationality, and in 18\% of cases its target was a certain political group (e.g. supporters or opponents of the current political regime in Russia). 13 messages had a certain social group as their target, poor or uneducated people were usually lashed out at. Geopolitical (usually based on nationality), confessional and pragmatic (critisizing the wat the message is written) aggression is much rarer.

In the course of the study linguistic features expressing aggression were also found. 12.5\% of all messages contain
some level of aggression, and there were various ways to express it. The most popular way to demonstrate aggression is
using distinct offensive words, especially obscene ones which are contained in 175 messages of 333 (53\%). Other
messages often contain common Russian words that do not usually carry any negative connotation (e.g.
``\textcyrillic{{\cyrsh}{\cyrk}{\cyro}{\cyrl}{\cyrsftsn}{\cyrn}{\cyri}{\cyrk}'' - ``schoolboy'' is used to offend a
communication partner.) Many messages transfer aggression by phraseological units containing profanity or not
(``{\cyrt}{\cyrery} {\cyri} {\cyrt}{\cyra}{\cyrk} {\cyrp}{\cyro} {\cyrzh}{\cyri}{\cyrz}{\cyrn}{\cyri}'' - ``you are an
ordinarily ...'' - always negative). Word building is quite often used to demonstrate aggression. In these messages
there are various types of word formation, e.g. root }compounding
(``\textcyrillic{{\cyrs}{\cyro}{\cyrc}{\cyri}{\cyro}{\cyrb}{\cyrl}*{\cyrd}{\cyrsftsn}'' - ``sociofag'') and blending
(``{\cyrk}{\cyrr}{\cyre}{\cyrm}{\cyrl}{\cyrya}{\cyrd}{\cyrsftsn}'' - ``kremlart'' - ``Kremlin'' + ``tart'',
``{\cyrp}{\cyrr}{\cyro}{\cyrp}{\cyra}{\cyrg}{\cyra}{\cyrn}{\cyrd}{\cyro}{\cyrn}'' - ''propacondom'' - ``propaganda'' +
``condom'' ) are quite popular. Affixation is also often used, e.g. ``{\cyrr}{\cyru}{\cyrs}{\cyrn}{\cyrya}'' -
``Russians'' + pejorative suffix.}

Aggression may be expressed not only via words and phrases, but via grammar, too. E.g. nominative sentences are used to
intensify the effect of the obscene word (``\textcyrillic{{\CYRD}*{\cyrl}{\cyrb}**{\cyrb}
{\cyrs}{\cyrch}{\cyri}{\cyrt}{\cyra}{\cyryu}{\cyrshch}{\cyri}{\cyrishrt} {\cyrch}{\cyrt}{\cyro}
{\cyrz}{\cyrn}{\cyra}{\cyrn}{\cyri}{\cyre} {\cyrs}{\cyrv}{\cyrya}{\cyrz}{\cyre}{\cyrishrt} {\cyrv}
{\cyrv}{\cyri}{\cyrt}{\cyrch}{\cyrh}{\cyra}{\cyru}{\cyrs}{\cyre} {\cyrv}{\cyrl}{\cyri}{\cyrya}{\cyre}{\cyrt}
{\cyrn}{\cyra} {\cyrk}{\cyro}{\cyrn}{\cyre}{\cyrch}{\cyrn}{\cyrery}{\cyrishrt}
{\cyrp}{\cyrr}{\cyro}{\cyrd}{\cyru}{\cyrk}{\cyrt}'' - ``Butthead thinking that knowledge of witch-house connections
influences the final product''). Imperatives are also broadly used ({\guillemotleft}{\cyri}{\cyrd}{\cyri}
{\cyrl}{\cyre}{\cyrch}{\cyri}{\cyrs}{\cyrsftsn}{\guillemotright} - ``get treated''). Sometimes only semantics of the
sentence shows the presence of aggression in the message (``intellect is not connected to the brain'').}

Obscene words are often used on anonymous forums not to express aggression but for other reasons. The following ways
of using them were discovered on 2ch.hk (according to Levin's classification [2]):
\begin{enumerate}
\item
obscene words as interjections
\item 
obscene words to express indifference
\item
obscene words as insertions carrying no emotional charge
\item {
Obscene words as a pronoun (e.g. may be used to address other people
(``\textcyrillic{{\CYRP}**{\cyri}{\cyrk}{\cyri},
{\cyro}{\cyrt}{\cyrv}{\cyre}{\cyrt}{\cyrsftsn}{\cyrt}{\cyre} {\cyrm}{\cyrn}{\cyre}, {\cyrch}{\cyrt}{\cyro}
{\cyrerev}{\cyrt}{\cyro} {\cyrz}{\cyra} {\cyrsh}{\cyrt}{\cyru}{\cyrk}{\cyra} {\cyrt}{\cyra}{\cyrk}{\cyra}{\cyrya}?'' --
``F**gots, answer me, what is this?'') -- this phrase was not considered to be an insult}}
\item {
Obscene words as verbs substitutes (e.g. just as the English word ``f*ck'' may convey almost any sense )}
\end{enumerate}
\section{Conclusion}
Anonymous imageboards are notorious for their aggressiveness and profanity use. This research has confirmed it.
According to data analysis, anonymity does not influence the percentage values of obscene words. Moreover, the
percentage values of obscene words is even higher for non-anonymous social network vk.com then for anonymous
imageboards (although this value may fluctuate). Thus, the percentage values of obscene lexis do not depend on the
anonymity of users as much as it might have seemed. \ However, these values depend on the peculiarity of the community
and the results obtained should be examined in further research. According to iichan.hk experience, moderation and
wordfilter may significantly decrease the percentage of obscene words and thus increase the communicational culture.
Sites having no moderation (2ch.hk, vk.com/mdk, 4chan.org) have a high level of aggression. However, negative impact of
moderation is yet unknown and should be researched.

It was also determined that obscene lexis does not always carry aggression. Thus, only 12.5\% messages carried signs of aggression and 8.2\% contained obscene lexis to express positive emotions. Aggression was usually expressed towards a communication partner. This is an indirect evidence of poor communication culture on the imageboards. Many messages contain political or ethnic aggression but it may change with the growth of the set.

These preliminary results show that there is no correlation between the degree of anonymity and level of aggression. However, further research of other sites
is needed. The results of this research will help to create a database for analysis of aggressiveness in various texts and may make us closer to automatic understanding of texts.
Further researches will be aimed at examining the dependency between anonymity and aggression for various languages
drawing a final conclusion and creating a software product for the aggression analysis.

{
\textbf{Acknowledgments. }The survey is being carried out with the support of the Russian Science Foundation (RSF) in
the framework of the project {\textnumero} 14-18-01059 at the Institute of Applied and Mathematical Linguistics of the
Moscow State Linguistic University (scientific head of the project is R. K. Potapova).}

\end{document}